# An unsupervised long short-term memory neural network for event detection in cell videos

Ha Tran Hong Phan[a], Ashnil Kumar[a], David Feng[a,d], Michael Fulham[b,c], Jinman Kim[a]

[a] School of Information Technologies, The University of Sydney

[b] Sydney Medical School, The University of Sydney

[c] Department of Molecular Imaging, Royal Prince Alfred Hospital

[d] Med-X Research Institute, Shanghai Jiao Tong University

**Abstract:** We propose an automatic unsupervised cell event detection and classification method, which expands convolutional Long Short-Term Memory (LSTM) neural networks, for cellular events in cell video sequences. Cells in images that are captured from various biomedical applications usually have different shapes and motility, which pose difficulties for the automated event detection in cell videos. Current methods to detect cellular events are based on supervised machine learning and rely on tedious manual annotation from investigators with specific expertise. So that our LSTM network could be trained in an unsupervised manner, we designed it with a branched structure where one branch learns the frequent, regular appearance and movements of objects and the second learns the stochastic events, which occur rarely and without warning in a cell video sequence. We tested our network on a publicly available dataset of densely packed stem cell phase-contrast microscopy images undergoing cell division. This dataset is considered to be more challenging that a dataset with sparse cells. We compared our method to several published supervised methods evaluated on the same dataset and to a supervised LSTM method with a similar design and configuration to our unsupervised method. We used an F1-score, which is a balanced measure for both precision and recall. Our results show that our unsupervised method has a higher or similar F1-score when compared to two fully supervised methods that are based on Hidden Conditional Random Fields (HCRF), and has comparable accuracy with the current best supervised HCRF-based method. Our method was generalizable as after being trained on one video it could be applied to videos where the cells were in different conditions. The accuracy of our unsupervised method approached that of its supervised counterpart.

## 1. Introduction

Advances in stem cell biology and pharmacology research rely heavily on discoveries of changes in cellular behaviour, especially in cells that respond to various stimuli introduced into cell cultures. Methods used to monitor the health and the rate of cell growth usually depend upon observing cells cultured in-vitro to detect cell divisions and cell deaths [1, 9]. There are a wide range of procedures with luminescent, colorimetric or fluorescent assays that highlight targeted components in captured images or videos. These imaging data are then visually assessed to learn essential information about the biomolecular processes inside cells. Drawbacks of these assays are that they are usually static and destructive and hence do not allow for long-term cell monitoring [2]. Phase-contrast time-lapse microscopy is a non-destructive imaging technique that enables in-vitro, long-term, continuous monitoring and analysis of cell populations. Such automated time-lapse systems offer enormous potential for pharmaceutical and stem cell engineering applications [3-6]. Current microscopy systems have high throughput acquisition protocols but the 'bottleneck' now is analysis of the large volume of cell image and video data. Automating the detection of cell events offers the opportunity for more rapid, robust and quantitative analysis of cell behaviour [7-8].

The cell dynamics of most interest include changes in cell shape, division and movement. These activities often appear to be random in nature and are governed by many stochastic elements in biological systems [46, 47]. Cells grown in slightly different conditions can exhibit marked changes in behaviour. Further, the cellular activities have varying spatiotemporal patterns due to cell events that occur at different stages of maturation [48]. These facets of cell behaviour make it extremely challenging to design optimal computerised image/video features and classifiers. An additional problem is the lack of labelled training data. Each new experimental setting may require manual data annotation by experts who have highly specialised skills in the specific research area being studied. The inherent risk with such an approach is the different levels of experience and knowledge that can then affect the ability to extract objective, reproducible quantitative information from large datasets. Hence, there is growing interest for innovative solutions for the detection of cell events in phase-contrast microscopy data. Since 2010, a number of methods for cell event detection have been reported. These include the Hidden Conditional Random Fields [14-18, 20] and the semi-Markov [19] methods for stem cell division event detection, and methods for generic cell event detection method [21, 22]. All these methods, however, require labelled data for training and so are limited in their ability to be widely applied to multiple cell types and events. Most cell division and cell death detection methods have an initial stage where potential event candidates are selected and then a machine learning method classifies the candidates into event and non-event groups [14-22]. They need extensive annotated data for training that can be problematic to acquire. An alternative approach is to track individual cells throughout the video and infer the cell division and cell death moments from the cell trajectories. Cell tracking itself is challenging, error-prone and requires careful parameter optimization [49, 50].

Deep neural networks (DNNs) allow feature extraction directly from in-domain data, thus enabling the learning of task-adapted feature representations [25]. Convolutional neural networks (CNN) excel in computer vision tasks because of their ability to extract hierarchically increasingly complex features, leading to a marked increase in image classification and detection accuracy in static images [23, 24, 31]. CNNs have excellent performance in visual feature extraction but, by themselves, they are not usually designed to detect patterns in video analysis. The long-short term memory (LSTM) method that is a variant of recurrent neural networks, learns the patterns in time-series data and is now the leading technique for natural language and video analyses [26-30, 32]. A neural network that combines feature extraction and dynamic changes over time then provides the best approach for cell video analysis. Recent advances in deep learning have shown promising evidence for the development of convolutional recurrent neural networks with CNN and LSTM components that can be trained in an unsupervised manner to learn the overall dynamics of objects in image sequences [35-37]. The possibility of unsupervised training of DNNs is highly relevant to cell video analytics. These methods are designed with a feed forward structure that passes a 3 dimensional tensor of a sequence of raw images through the neural network and output another 3 dimensional tensor of the predicted image sequence but such approaches are not able to solve problems where the events are unpredictable [36] such as events determined by underlying stochastic biomolecular interactions. Hence, current methods are not directly applicable to cell event detection, in which the timing and location of each event must be determined simultaneously.

Current methods for cell division detection are supervised and directly search for the cell division events and ignore other objects and activities in an image sequence [14-22]. The ideal method is one that performs both detection for stochastic events and pattern learning for common object dynamics. It should also reduce or remove the need for the generation of labels through use of semi-supervised or unsupervised methods, as suggested in a recent review by Liu et al. [51]. In this research we propose an end-to-end trainable convolutional LSTM (ConvLSTM) model, where the convolutional component is integrated into LSTM to capture video spatiotemporal structures. Our method extends prior work in ConvLSTMs by Shi et al [35] through a branched structure that reads both the target sequence and the frames preceding it, then merges the learned states of the cell population to reconstruct the target sequence, while simultaneously extracting information of potential events. We applied it to phase-contrast microscopy videos of a publicly available dataset which was used by Huh et al, which is regarded as state-of-the-art work [18]. Our contributions are as follows, we introduce: a) a new convolutional LSTM network that addresses the partially stochastic nature of cell activities with a sequence-to-sequence learning framework; b) unsupervised end-to-end training using spatiotemporal convolutional filters and max-pooling to learn condensed representations of potential video cell events and c) unsupervised detection of cell divisions in densely packed cell videos.

## 2. Related work

Early research in the detection of cell events, from 2005 to 2009, relied on cell tracking methods that created trajectories of cells by matching objects frame-by-frame, and then extrapolated the moments of cell events by analysing those trajectories [10-13]. Variability in cell appearance, lack of predictability of cell movement and similarities between target and adjacent cells resulted in inaccurate detection. Later research directly targeted the events of interest through Support Vector Machine (SVM) and Hidden Conditional Random Fields (HCRFs) methods [14-20]. HCRFs proved superior to SVM methods as they were better able to model the temporal location of cell divisions and the change in the cells' appearance over time. A series of HCRF models have been proposed, advancing from only patch sequence classification [16] to identifying the timing of cell splitting moments in a sequence [17, 18]. The HCRFs methods, however, rely on hand-crafted features that require considerable effort to select and optimise for different application domains when compared with features learned with deep learning [23]. Thus HCRFs methods tend to be application-specific, e.g. to detect cell divisions and cell deaths, and are often optimized for a particular cell type only. The problem domain is the detection of cell divisions amongst a dense collection of cells [18, 51] and for this HCRFs models are the current benchmark.

Recently new methods that require less training data and offer more flexibility have been developed. The autoregression model reported by Kandemir et al. [21] uses a small set of event-free training data for generic cell event detection. The collaborative multi-output Gaussian process and active-learning requires only 3% of the annotated data [22]. Nevertheless, these methods still require manual annotation of the training data. Deep learning approaches to detect cell divisions, such as the 2D CNN that detects the "figure-8" shape of dividing cells in phase-contrast microscopy, were reported by Shkolyer et al. [43]. A 3D CNN for a volumetric region was used by Nie et al. [44]. These methods replace a step in the pipeline of other traditional methods for cell division detection, for instance as a classifier or a feature extractor before applying standard classifiers [43, 45]. These methods require careful algorithm selection for the pre- and post-processing steps, are supervised models and must be trained with large annotated datasets. As such, they are not feasible in real-world scenarios where there are few or no labels. Advances in deep learning have now made unsupervised training a practical possibility. Recurrent neural networks (RNNs) are end-to-end trainable models. They can be trained in a supervised or unsupervised manner, retain memory of seen data and can be applied to tasks involving time-series imaging data [33-39]. Work by a number of researchers used encoder-decoder LSTM to learn video representations [33, 34]. They also extended the fully-connected LSTM to have convolutional structures that better captured spatiotemporal correlations in an image sequence [35-37]. Other researchers used multiple LSTMs to classify different components of video events [38, 39]. These methods require thorough annotations for the events or action labels for each person tracklet, which are difficult to replicate in biomedical research.

## 3. Theoretical background

LSTM units were originally proposed by Hochreiter and Schmidhuber [29]. They consist of different memory cells that store information for a short time, and gating components that control the flow of the content being stored. We used the formulation of LSTM reported by Graves et al. [26].Figure 1 shows an LSTM unit (based on the formulation of Graves et al [26]) with a memory unit, holding a state $c_t$ at time t, which acts as an accumulator of the state information. Three self-parameterized controlling sigmoidal gates, including input gate $i_t$, forget gate $f_t$ and output gate $o_t$, govern the access to this memory unit (Fig. 1). For every new frame, two external sources and an internal source are received at the four terminals (the input and the three gates) of a LSTM unit. The external sources include the input frame $x_t$ and the previous hidden states of LSTM units at the same layer $h_{t-1}$, while the internal source is the memory cell state $c_{t-1}$. The memory cell will accumulate new information if the input gate is activated, or remove the past memory cell status $c_{t-1}$ if the forget gate $f_t$ is triggered. The output gate $o_t$ controls whether the state $c_t$ will be propagated to the final hidden state ht. The gates are activated by passing the accumulated inputs, along with a bias, through a sigmoid function. The memory cell state is computed using the previous memory cell state $c_{t-1}$ after passing through the forget gate, and the inputs at the input terminal after getting through the tanh non-linearity and input gate. The updated memory cell state is then passed through tanh non-linearity and the output gate to get the final output, the hidden state $h_t$. This gating mechanism allows for the back-propagated gradient to be trapped in the memory cell and prevented from vanishing or exploding quickly, thus enabling learning long range dependencies.

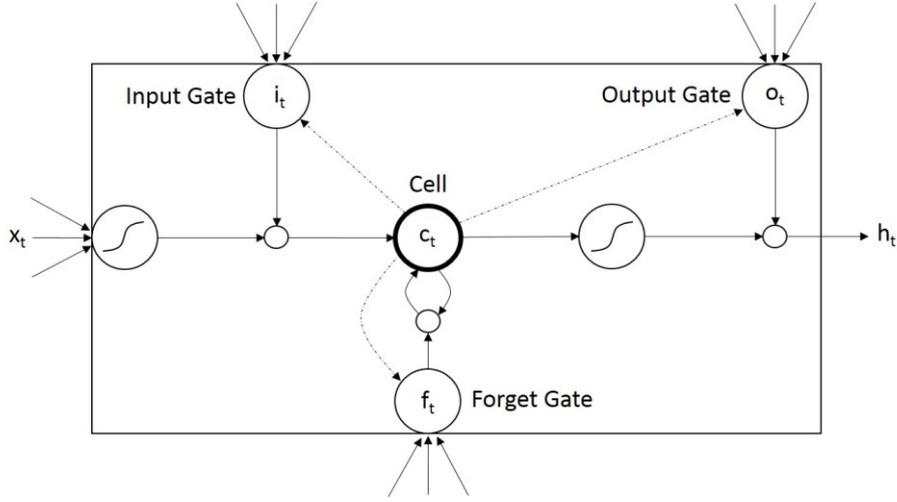

Figure 1: Structure of an LSTM unit.

The above LSTM structure utilizes full connections in input-to-state and state-to-state transitions that ignore spatial information, which is essential in modeling the regularities in image sequences. We chose the convolutional LSTM (ConvLSTM), based on the work of Shi et al. [35]. This LSTM preserves spatial and temporal information for modelling long-range dependencies and better encodes the spatio-temporal information of cell image sequences. The major advance in ConvLSTM is the use of 3D tensors for all inputs $x_t$, cell outputs $c_t$, hidden states $h_t$ and gates $i_t$, $f_t$, $o_t$ in which the last two dimensions encode the spatial dimensions. The Hadamard products between the weights and inputs $x_t$ or hidden states $h_t$ are then replaced with convolution operators:

$i_t = \sigma(W_{xi} * x_t + W_{hi} * h_{t-1} + W_{ci} \circ c_{t-1} + b_i)$

$f_t = \sigma(W_{xf} * x_t + W_{hf} * h_{t-1} + W_{cf} \circ c_{t-1} + b_f)$

$c_t = f_t \circ c_{t-1} + i_t \circ \tanh(W_{xc} * x_t + W_{hc} * h_{t-1} + b_c)$

$o_t = \sigma(W_{xo} * x_t + W_{ho} * h_{t-1} + W_{co} \circ c_t + b_o)$

$h_t = o_t \circ \tanh(c_t)$

where ($\circ$) denotes the Hadamard operator and (*) denotes the convolution operator. As the convolution operation is performed on the inputs and states, padding is required to ensure that the states have the same dimensions as the inputs.

## 4. Our LSTM approach

For our LSTM method, let $x_t \in [0, 1]^{M \times M}$ represent a frame of size MxM, and $y_t \in [0, 1]^{n \times M \times M}$ represent the possible events of n classes in that frame at time *t*. The problem therefore is to solve $P(y_{t:t+k} | x_{t:t+k})$, the probability of the events occurring at time steps *t* to *t+k*, given the frames $x_{t:t+k}$. In our unsupervised approach, the only input for the model is the cell imaging sequence; there is no ground truth to train a model to produce $y_{t:t+k}$ directly. Given the range and variability of cell differences and activities the problem is unlikely to be solved if given only the frames before $x_{t:t+k}$. Therefore, we extract the information about possible events in a short cell imaging sequence and reconstruct the sequence by combining the learnt cellular dynamics from the frames preceding the sequence with the extracted event information. We use the observed frames $x_{1:t-1}$ to learn the dynamics of cell behaviour and appearance until time step *t-1* and the target frames $x_{t:t+k}$ to extract event information to model the complete states of the cell population at time steps *t* to *t+k* and reconstruct $x_{t:t+k}$.

To solve this problem, we model the dynamics of cellular activities, $P(h_t|x_{1:t-1})$, with $h_t$ being the latent state that captures the complete dynamics of objects up to time *t-1*, and then predict the future frames $x_{t:t+k}$ with the probability $P(x_{t:t+k}|h_t)$. For example, cells tend to follow their current motion direction and appearance, which happen from time 1 to *t-1*, and carry on at time steps *t* to *t+k*. To capture the complete state of the cell population, the problem is formulated with the joint probability density:

$P(x_{t:t+k} | x_{1:t-1}, y_{t:t+k}) P(y_{t:t+k} | x_{t:t+k})$

$= P(x_{t:t+k} | h_{t-1}, y_{t:t+k}) P(h_{t-1} | x_{1:t-1}) P(y_{t:t+k} | x_{t:t+k})$

(Eq. 1)

where

- $P(h_{t-1} | x_{1:t-1})$ models the hidden state transition probability capturing the dynamics of the cell population up to time step *t-1*;
- $P(y_{t:t+k} | x_{t:t+k})$ describes the event detection at time steps *t* to *t+k*;
- $P(x_{t:t+k} | h_{t-1}, y_{t:t+k})$ models the reconstruction process that extrapolates the future frames from the learnt dynamics $h_{t-1}$, with additional information of potential events $y_{t:t+k}$.

### 4.1 Encoding-Reconstruction Structure

Since detected events are only biologically meaningful at the level of individual cells, the model should output only one class of events for each region corresponding to the area of an average cell. We integrated downscaling max-pooling and softmax operations into our model to produce spatially defined classes of events. Figure 2 depicts our ConvLSTM model. There are 3 parts (E, R, Ev). Each part learns a function to model one of the three probabilities in equation (Eq. 1). The Encoding branch (E) learns the movements of the cell population in frames before the target sequence. The Event Detection branch (Ev) generates the cell event detection and classification for the target sequence. The Reconstruction part (R) combines the outputs of E and Ev to reconstruct the target sequence $x_{t:t+k}$, as formulated in (Eq. 1).

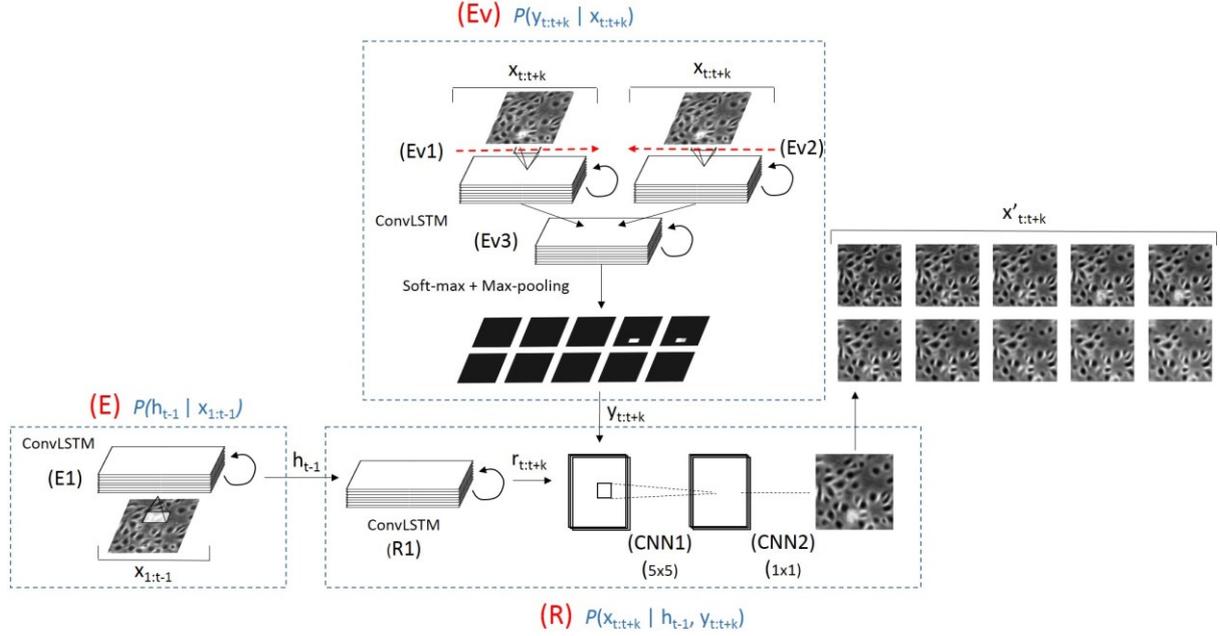

Figure 2: Outline of our ConvLSTM model.

The encoding ConvLSTM, branch E, labelled E in (Fig. 2), models $P(h_{t-1} \mid x_{1:t-1})$ to compress the input sequence $x_{1:t-1}$ into a hidden state $h_{t-1}$, which contains the information about the dynamics of the cell population. This branch consists of a ConvLSTM (E1) that takes as input an image sequence, $x_{1:t-1}$, which contains the frames prior to the target sequence that we want to detect cell events from. E1 then extract visual features and temporal patterns of the cell population in $x_{1:t-1}$. The last states and cell outputs of E1 are then copied as the initial states and cell outputs of the subsequent decoding ConvLSTM (R), which unfolds the hidden state $h_{t-1}$ and uses it to model $P(x_{t:t+k} \mid h_{t-1}, y_{t:t+k})$ (Fig. 2).

The second branch of the model, branch Ev, detects events by learning the probability function $P(y_{t:t+k} \mid x_{t:t+k})$. Firstly, each of the two ConvLSTMs (Ev1, Ev2) reads the sequence $x_{t:t+k}$ in either the forward (*t* to *t+k*) or backward (*t+k* to *t*) direction (Fig. 2). As cell events, such as cell divisions, happen in multiple phases and each phase possesses different characteristics in visual appearance, these two ConvLSTMs are used to ensure that both the past and future changes in a cell in each single frame are captured. Another ConvLSTM (Ev3) then combines the hidden states generated from both directions of sequence reading to learn the dynamics of all objects in $x_{t:t+k}$. The final hidden state tensor produced from Ev3 is then used for the event detection step (Fig. 2).

As we expect to detect and classify the potential events into distinct classes, the event detection step will output a 3D tensor in which the first dimension indicates the classes of events (16 in our model) and the last two dimensions contain the spatial information (64x64). A problem with event classification is that, with n classes of events to detect and each pixel is to be assigned to at most one event class, the convolutional neural network can learn to assign adjacent pixels to different classes and later combine them together to implicitly form much larger than n number of event classes. We would like to keep n small so that only the essential cellular events are to be detected.

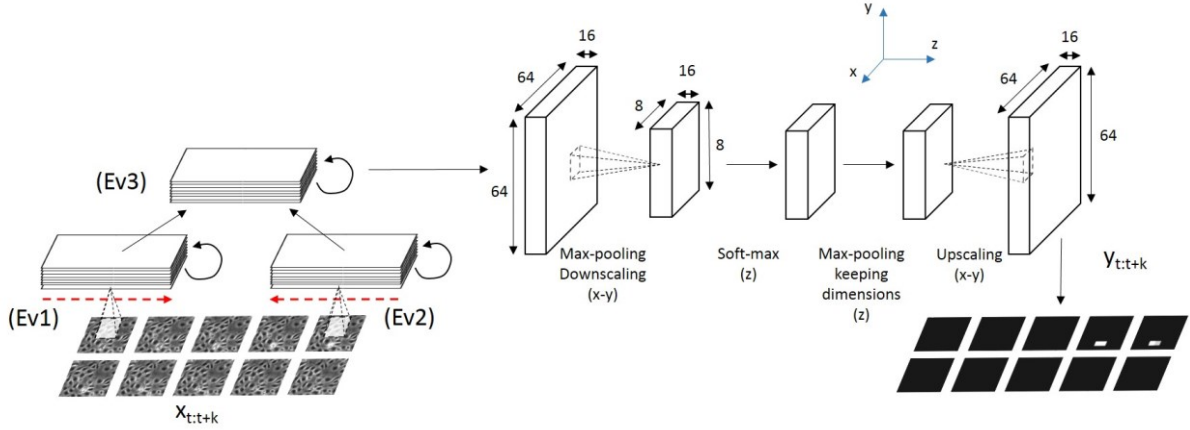

Figure 3: Softmax and max-pooling for event detection in our ConvLSTM. The dimensions of the tensor being passed through this part of the model are displayed to illustrate the operation of the max-pooling and up-scaling layers.

To avoid this problem, we keep a low spatial resolution for event detection, of 8x8 pixel squares which are equivalent to the size of an average cell, so that for each cell, only one event class is recorded. This turns an image of size 64x64 to a grid of size 8x8. We use max-pooling and softmax as shown in Figure 3 to extract only one event type for each location in the grid. The process consists of the following steps:

- The hidden state tensor is down-sampled by a factor of 8 with max-pooling, with no overlap on pooling regions (*x-y* dimensions).
- The softmax function is applied across the classes for each spot in the grid; this step turns all values into non-negative numbers and generates a categorical distribution between the possible classes (*z* dimension).
- Another max-pooling step sets non-maximum values among the classes to zero, thus only the class with the highest activation value remains activated (*z* dimension).
- The 3D tensor is then scaled back to its original shape of 64x64 (*x-y* dimensions).

This entire process can be considered similar to grouping similar events, and each group can be thought of as a class. An event class is determined for each 8x8 pixel grid, which covers the area of an average cell.

To combine the outputs of E and Ev to reconstruct the target sequence $x_{t:t+k}$, the final component R, which learns the probability, $P(x_{t:t+k} | h_{t-1}, y_{t:t+k})$, can be modelled with a ConvLSTM (R1) and convolutional layers (CNN). R1 unfolds the hidden state tensor $h_{t-1}$, which is a compression of the previous sequence from E, into a prediction of the target sequence ($r_{t:t+k}$). However, this prediction does not account for the stochastic influences on cellular events. The event detection result, $y_{t:t+k}$, is thus concatenated with the output $r_{t:t+k}$ of R1 to provide complete information for the convolutional layers (CNN1 and CNN2) to regenerate the target sequence (Fig. 2). The first convolutional layer (CNN1), which has filter kernels of size 5x5 to cover the area around a pixel, takes as input the combination of the event classes, $y_{t:t+k}$, and mechanics of objects encoded in the output of R1, $r_{t:t+k}$. The next convolutional layer (CNN2) with filter kernels of size 1x1 produces the reconstructed images of the target sequence, $x_{t:t+k}$.

## 4.2 Post-processing step for cell division detection

The output of the model indicates the areas where events of different classes occur. As such, a minimal post-processing step is required to determine the centroid pixels of the events' locations. As the outputs are generated in an unsupervised manner, they contain clusters of events. Therefore, some external heuristic knowledge is needed to specify the event of interest; cell division detection requires the selection of the class of events, out of the n classes outputted, that contain cell divisions. After obtaining the map of detected cell division events, our post-processing stage groups pixels with positive activation values into patch sequences of events. Then, the centroid pixel of the cell division event is located based on the knowledge that cells shrink during cell divisions, which leads to a peak in brightness in phase-contrast microscopy images. For these reasons, the cell division events' locations were determined by finding the pixel with the highest average increase in intensity after some time $t$, set to be 2 frames which is about one phase in the cell division process according to our videos' frame rate, over a circular area $a$ with a radius of 5 pixels, which corresponds to the size of cells undergoing divisions.

# 5. Implementation

We used the Theano framework, an open-source library for deep learning algorithms, with support for the use of GPUs to implement our method. We trained it on an NVIDIA 12GB Titan X (Maxwell architecture). We pre-processed the sequences by linearly rescaling image pixel values so that the final range was between 0 and 1. Each original frame of size 1040x1392 pixels contained a large amount of cells and as such has a large memory footprint. We reduced this by dividing the sequence into smaller subsequences of 15 frames, each with 256x256 pixels, by using a moving window method with spatial steps of 128 pixels and temporal steps of 1 frame. These frames were then downscaled by a factor of 4 into frames of 64x64 pixels. We used a data augmentation method to reduce the likelihood of overfitting during training [45]. The augmented samples were generated by flipping each subsequence horizontally and vertically, and rotation by 90, 180, and 270 degrees. Our architecture's hyper-parameters, including state numbers, kernel and max-pooling size for each layer are summarized in Table A.1 in the Appendix. Each ConvLSTM layer had 32 hidden states and all the input-to-state and state-to-state kernels were 5x5. We empirically determined a small set of architecture hyper-parameters based on previous work [35]. By performing zero-padding on the states, we assume no prior knowledge about the biological cells that are outside of the boundary of the video frame. The model was trained by minimizing the cross-entropy loss between $x_{t:t+k}$ and $x'_{t:t+k}$ using RMSProp [40] and back-propagation through time [41], with a learning rate of $10^{-3}$ and a decay rate of 0.9 [40, 41]. Biases were initialized to zero and weights were generated using Xavier initialization [42]. We trained our neural network over 100 epochs.

# 6. Methods

## 6.1 Dataset

We used a publically available dataset of densely packed stem cell phase-contrast microscopy images to train and evaluate our approach. This dataset is considered to be more difficult than other datasets with sparsely located cells [51]. We used 5 videos of bovine aortic endothelial cells (BAEC)[1] that were also used by Huh et al. [18]. Each video contains 210 images. The videos begin with cells separated on the left and right with an empty area in between, and end with cells completely filling the frame after cell divisions. There were two control videos without any added drug (F0001, F0002), and three videos where a drug, Latrunculin B, was added (F0003: 1nM, F0004: 10nM, F0005: 100nM). Cells moved more slowly and had fewer cell divisions with higher drug doses. The dataset includes the ground truth of manual annotations of cell divisions and the counts as shown in Table A.2 in the Appendix. We followed the standard true position detection definition where a birth event, where two daughter cells are completely separated from each other, is correctly detected if it is spatially within 10 pixels and temporally within 1 or 3 frames from the manual annotation [18].

---

[1] Dataset source: http://celltracking.rit.albany.edu/index.php

## 6.2 Experiments

We compared the ability of our method to detect cell divisions to standard HCRF, Event Detection Conditional Random Field (EDCRF) and Two-Labeled Hidden Conditional Random Field (TL-HCRF) methods [17-18]. We used a similar approach as these methods and trained our models on the first 100 frames and tested it on the last 100 frames. Cell division events were evaluated in terms of precision and recall and F1-score, which is a balanced measure for both precision and recall, given the timing errors of birth event being 1 or 3 frames. The same training procedure and evaluation protocol was applied for the supervised LSTM model. To study the performance of the event detection branch of our model when cell division annotations are available, we designed a model built from the ConvLSTM layers from the Event Detection (EV) branch and the convolutional layers from the Reconstruction (R) part of the unsupervised model. This model can be trained in a supervised manner with the target containing the spatial and temporal information of cell divisions in the sequence (Fig. 4). For each frame, a small square of size 7x7 and a larger square of size 20x20 centring at each annotated pixel were assigned a value of 1.0 and 0.6, respectively, while the other pixels' values were set at 0.1. This provides incentives for the model to learn patterns around a cell event (20x20), and also to learn to focus on the central pixels (7x7) at the splitting area of the dividing cell. The output of the model is post-processed to find the detected pixel which is the centroid of a group of adjacent pixels of values above a threshold of 0.7.

To thoroughly investigate the performance of our models, we benchmarked our models against published supervised methods, and studied their generalizability over videos of different characteristics. We trained our models on one video (F0001) and tested it on the other 4 videos (F0002-F0005) from the same dataset. To further investigate the behaviour of our models, we also studied the frequency distribution of birth event timing errors of the cell division detection for all 5 videos.

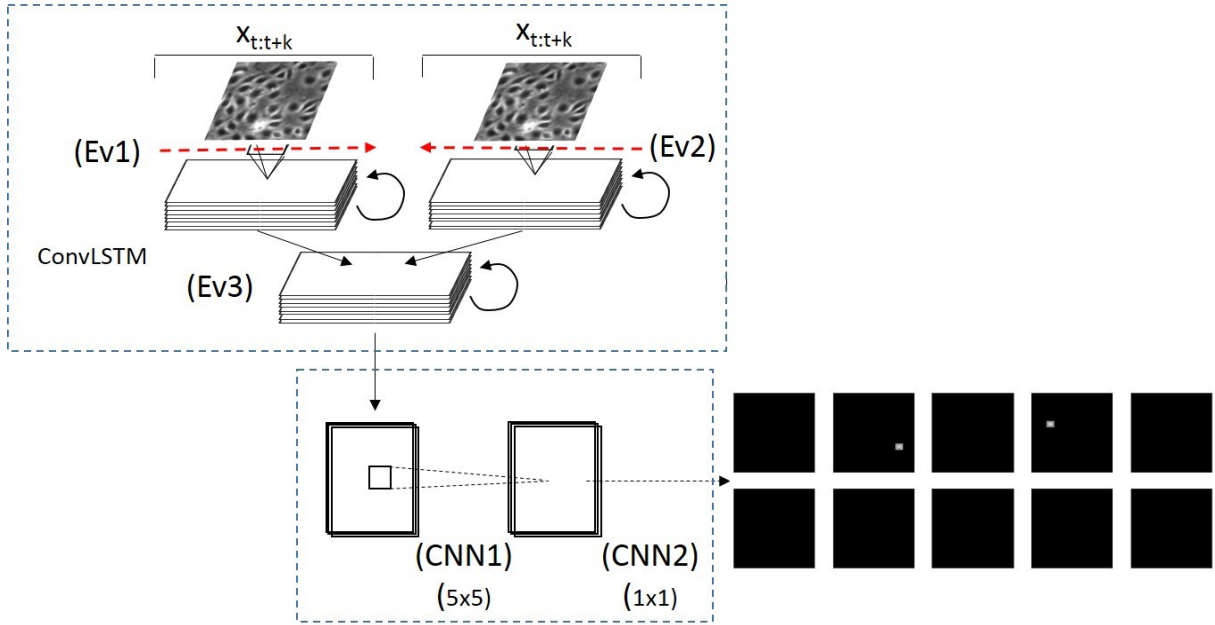

Figure 4: Supervised LSTM model. This model is comprised of the Event Detection (EV) branch, with the softmax and max-pooling layers removed, and the convolutional layers of the Reconstruction (R) part of the unsupervised LSTM model.

We also conducted experiments to investigate the behaviour of our model during the unsupervised training process, including how quickly the model learns to reconstruct the target sequence, the evolution of types of events that are chosen throughout the training process, and the convergence of the event detection. Each training epoch we extracted and visualized the outputs of:

- the last CNN layer to examine the model's progress in learning to reconstruct the target sequence
- the softmax and max-pooling layers to examine the evolution of the visual and temporal patterns that the model considers to be a potential event, and the convergence of the final event detection

### 6.3 Results

#### 6.3.1 Detection of cell divisions

In Table 1 we show the comparison of the performance of our unsupervised approach with supervised methods, including our supervised LSTM approach, standard HCRF, Event Detection Conditional Random Field (EDCRF) and Two-Labeled Hidden Conditional Random Field (TL-HCRF) [17-18]. In terms of F1-score, as expected the supervised LSTM model achieves the best accuracy, while our unsupervised LSTM model approaches the accuracy of the best HCRF-based model, the TL-HCRF, within 0.03 in score. In Table 2 we show the performance of our models over videos of cells being cultured under various conditions. Both LSTM models perform consistently across 5 videos and the unsupervised LSTM model approaches its supervised counterpart, within 0.06 in F1-score, when the threshold of 3 frames applied. In addition, we also calculated the frequency distribution of birth event timing errors of the cell division detection for all 5 videos in Fig. 5. The supervised LSTM method achieves higher temporal accuracy than the unsupervised LSTM method, which is expected as annotation narrows down the search space for the neural network.

Table 1: The cell division detection accuracy results of four methods in terms of precision, recall, F1-score when applying the threshold of 1 frame (th=1) in distance from the annotation instead of 3 frame (th=3).

|  | th=1 | | | | | th=3 | | | | |
| --- | --- | --- | --- | --- | --- | --- | --- | --- | --- | --- |
|  | Unsupervised LSTM | Supervised LSTM | HCRF | EDCRF | TL-HCRF | Unsupervised LSTM | Supervised LSTM | HCRF | EDCRF | TL-HCRF |
| Precision | 0.767 | **0.943** | 0.622 | 0.641 | 0.796 | 0.856 | **0.950** | 0.695 | 0.718 | 0.847 |
| Recall | 0.578 | 0.636 | 0.604 | **0.654** | 0.650 | 0.644 | 0.640 | 0.675 | **0.733** | 0.692 |
| F1-score | 0.659 | **0.759** | 0.613 | 0.647 | 0.716 | 0.735 | **0.765** | 0.685 | 0.726 | 0.761 |

Table 2: The cell division detection accuracy results of our models for 5 videos in terms of precision, recall, and F1-score when the allowed timing errors of birth event are 1 and 3 frames from the annotation.

|  |  | Model trained with video F0001 | | | | | | | | | |
| --- | --- | --- | --- | --- | --- | --- | --- | --- | --- | --- | --- |
|  |  | th=1 | | | | | th=3 | | | | |
|  |  | F0001 | F0002 | F0003 | F0004 | F0005 | F0001 | F0002 | F0003 | F0004 | F0005 |
| Unsupervised LSTM | **F1-score** (Precision/Recall) | **0.659** (0.767/0.578) | **0.605** (0.622/0.588) | **0.690** (0.713/0.669) | **0.689** (0.763/0.628) | **0.544** (0.589/0.506) | **0.735** (0.856/0.644) | **0.768** (0.791/0.747) | **0.762** (0.787/0.738) | **0.742** (0.822/0.677) | **0.759** (0.822/0.706) |
| Supervised LSTM | **F1-score** (Precision/Recall) | **0.759** (0.943/0.636) | **0.765** (0.852/0.693) | **0.819** (0.942/0.724) | **0.823** (0.983/0.707) | **0.795** (0.935/0.690) | **0.765** (0.950/0.640) | **0.813** (0.906/0.738) | **0.819** (0.942/0.724) | **0.809** (0.966/0.695) | **0.822** (0.968/0.714) |

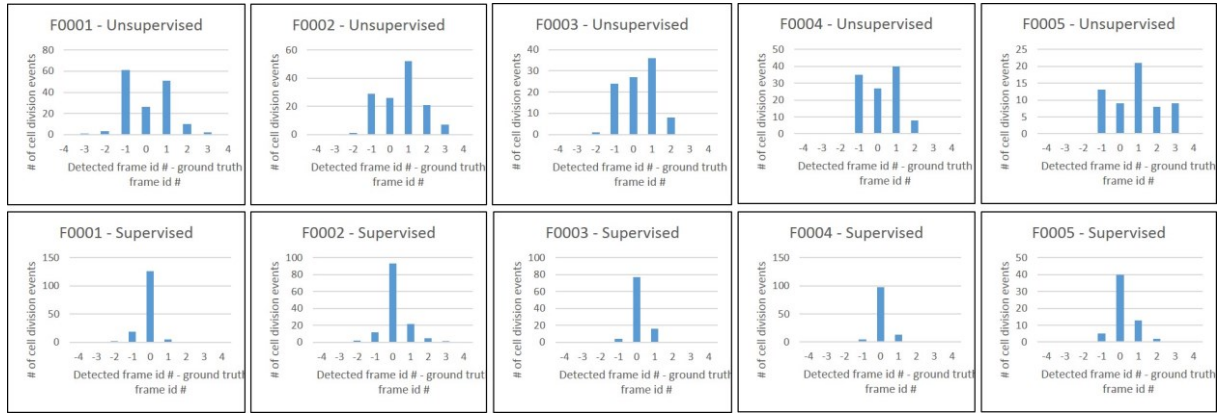

Figure 5: Temporal localization precision of cell division detection by our models for 5 different videos.

### 6.3.2 Model behaviour

In Fig. 6 and Fig. 7 we show the outputs of the model at selected epochs throughout the training process. The outputs are densely sampled in the early epochs (1, 10, 20, 30, and 40) because most of the essential learning about the characteristics of the imaging data happens in this epoch range, while the later training of the model, from between epochs 40 to 100, focuses on the refinement of the learned features and patterns.

In Fig. 6 we show the outputs from the last CNN layer of the model. After the first few epochs, the model learns the shape and the various movement patterns of cells as well as tracks their position in the future frames. From epoch 1 to 20, the reconstructed frames at the end of the sequence are blurrier than the first frames. From epoch 30, the network produces the complete reconstructed sequence with cell events, such as cell divisions marked with red circles in Fig. 6, at frames 2-5 and 6-7.

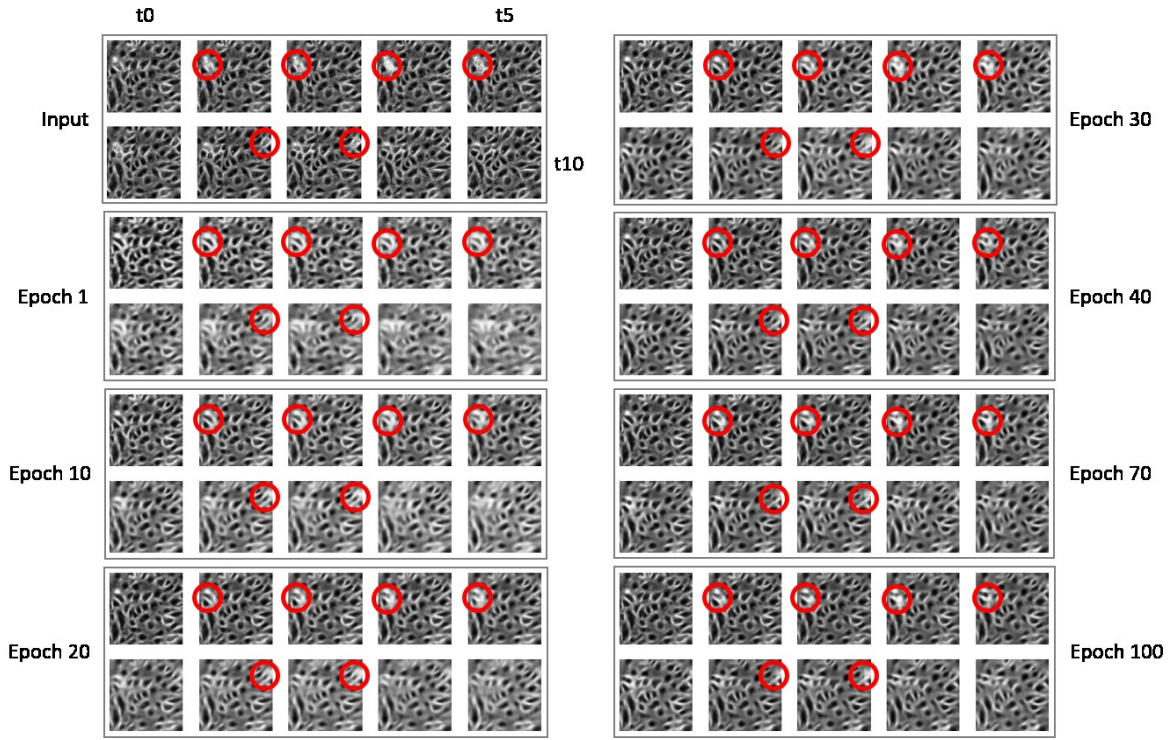

Figure 6: Reconstructed frames, which are the outputs from the last layer of the neural network. The original input target sequence and the outputs sampled at 7 different epochs contain 10 small images in 10 time points in the target sequence. The order in time of the small images starts from left to right and top to bottom. The red circles indicate the cell division events; there are two cells dividing here at frames 2-5 and 6-7.

In Fig. 7 we display the output of the softmax and max-pooling layers at epoch 100, which detects the temporal and spatial location of cell division events.

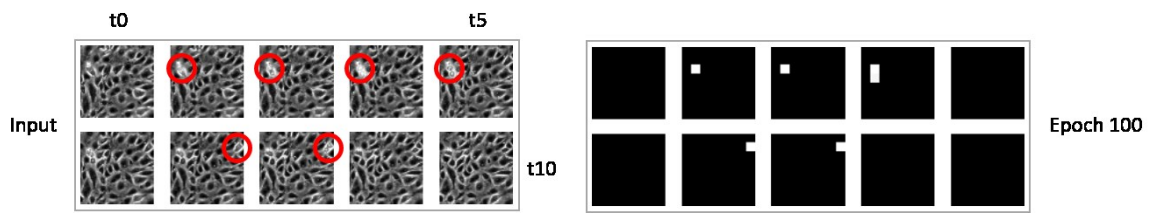

Figure 7: Event detection which is the output of the Event detection ConvLSTMs and softmax and max-pooling layers at epoch 100. The red circles indicate the cell division events; there are two cells dividing here at frames 2-5 and 6-7.

The events that were captured by the Event Detection ConvLSTMs (Ev) differ in both visual features and temporal dynamics. In Fig. 8, 5 out of 16 types of events are shown that are outputted by Ev at epoch 100; to display the context around the events, we have expanded the regions where events occur (like in Fig. 7) by a few pixels. Because cell divisions are the events of interest in this dataset, the class of this type of event is shown in Fig. 8(b), together with 4 other classes that have detectable temporal patterns but are not cell divisions. From our visual inspection of the model's outputs, cell division ends up in its own unique group.

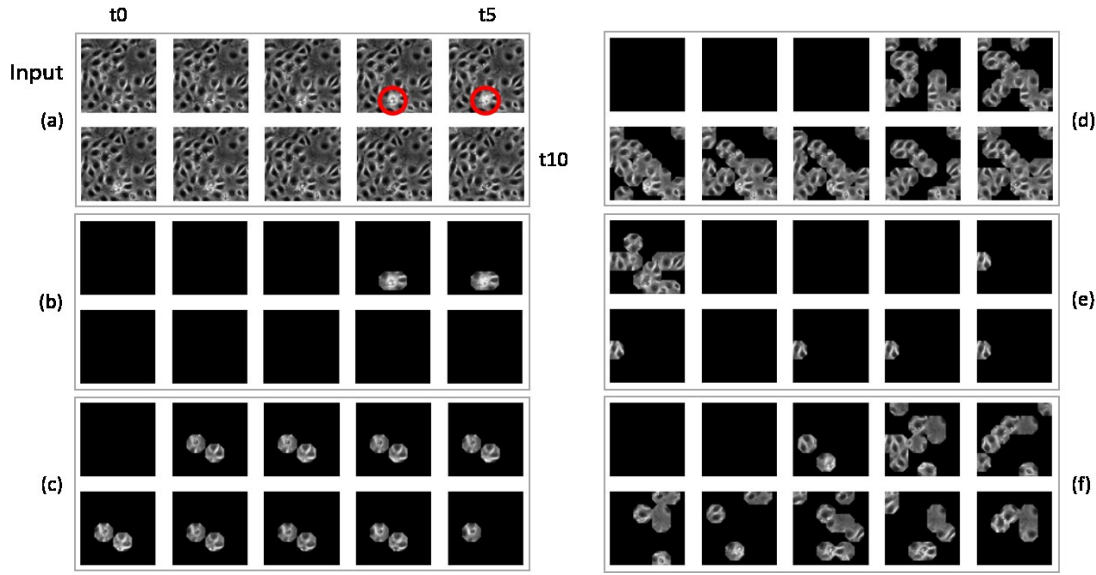

Figure 8: Demonstration of some of the events detected by the neural network. 5 classes of events are displayed here (b-f). The class (b) is associated with cell division events.

## 7. Discussion

Our results show that our unsupervised approach is more accurate than or has similar performance in the detection of cell divisions when compared to the benchmark supervised methods (Table 1). Our approach is also able to group biologically meaningful events into groups. In addition, our experiments showed that it has a consistent accuracy performance across videos with different characteristics (Table 2).

The F1-score in Table 1 shows that our unsupervised deep learning method consistently achieves higher scores than HCRF, comparable scores with EDCRF, and approaches TL-HCRF with a score that is 0.026 lower when the threshold is 3 frames, and 0.057 lower when the stricter condition of 1 frame timing error is applied. Our unsupervised approach demonstrated a better overall scores than HCRF and EDCRF because of the highly accurate spatial and temporal locations of cell events provided by the event detection (Ev) (Fig. 2) that narrows down areas where cell division would occur. HCRF only uses one label variable to model the occurrence of cell divisions and as such lacks the ability to represent birth event timing. EDCRF was designed to model the timing of birth event and as such have an additional label to represent the event timing [17], and thus able to achieve comparable accuracy with our unsupervised model. TL-HCRF is an upgrade of EDCRF that leverages the information on candidate birth event timing to detect the cell division events and is highly optimized for cell division detection [18]. This explains the smaller drop in F1-score of TL-HCRF as compared to the other HCRF models and our unsupervised LSTM when applying the stricter timing error condition. When using a threshold of 3 frames, our unsupervised approach scores a higher precision rate than the supervised HCRF models. This indicates that our model possesses high selectivity in the classification of potential events, leading to lower false positive rate in cell division detection. Many models' outputs are further analyzed with post-processing steps. Post-processing requires a high precision rate, especially in unsupervised training when the classes of events to be learned are unknown, such that it will not be overwhelmed by noise from false positives. In this aspect, our unsupervised model shows superiority over the HCRF models that have lower precision despite being trained with labels and optimized for just one event class (cell division).

The benchmark HCRF methods are multi-stage pipelines with a candidate patch sequence construction to skew the distribution of training data for the HCRF models by balancing the ratio of negative and positive samples. They rely on a supervised step with support vector machine (SVM) to reduce the search space for cell division events, followed by a second supervised step with a probabilistic model derived from standard HCRF to determine whether each candidate sequence actually contain a cell division event. These models utilize selected hand-designed features and heavily supervised training processes that are optimized separately. Our end-to-end trainable neural networks scan the entire raw imaging input data, without altering the training data distribution, and directly outputs the map of detected events without the need to reduce search space. Furthermore, the HCRF models are specifically designed for cell division detection with structures containing sub-labels that take into account the temporal location and timing of the cell splitting moment in addition to the visual features' changes, while our unsupervised model was designed for event detection in general without a dedicated component for capturing the patterns of changes in cell division progress.

The supervised LSTM model, which has a similar structure with the event detection branch of the unsupervised model, scores the highest precision rates in both evaluation conditions, with over 0.1 difference when compared with TL-HCRF. The lower recall rates by both of the LSTM models indicates that our deep neural networks have a tendency to favor precision at the expense of recall. This could be due to the loss function depending on the reconstruction of the input that puts too much focus on reconstructing parts of the sequence that are not related to cell events and overlooks transient cell divisions of cells of small sizes. An advantage with high precision is that low false negative rate is always a desired outcome in cell research, especially in studies when further analysis of detected events is required after the detection step.

Our model has a consistent accuracy performance across videos of different characteristics (Table 2). With a threshold of 3 frames (th=3), the F1-scores of the unsupervised LSTM model are within a range from 0.735 to 0.768, which are higher than the F1-scores of supervised HCRF models under the same evaluation conditions (Table 1). Under the stricter condition for cell division detection with a threshold of 1 frame (th=1), the F1-score of the unsupervised LSTM model (Table 2) drops below 0.6 only for video F0005, which contains cells with the highest drug dose. This can be explained by the frequency distribution of birth event timing errors of the cell division detection in all 5 videos (Fig. 5). As seen in Fig. 5, the case of video F0005 has the highest variance in timing errors, which is because it is the video with the most significant difference in cell appearance and cell event processes that are affected by the effect of the highest drug dose. However, despite these differences that causes a lower temporal accuracy, our model still maintains a stable spatial accuracy, proven by a comparable F1-score with that of other videos when using a threshold of 3 frame instead of 1 frame.

In Fig. 5 we show that the temporal localization achieved by our supervised model is more accurate than the unsupervised model. When using a strict timing criterion with a threshold of 1 frame, the unsupervised one gained a F1-score of 0.1 to 0.251 lower; however, with a threshold of 3 frames, it maintained a F1-score that was within 0.03 to 0.06 of the supervised LSTM, as in Table 2. These results demonstrate that the combination of learning of both regular objects' dynamics and stochastic objects' events in a video significantly enhances a model's capability in detecting events. Learning regular dynamics allows capturing of common visual changes while learning stochastic events allows us to model changes influenced by the underlying biological processes of essential cell events.

The behavior of our model can be analyzed using Fig. 6-8. Our results show that our model is capable of learning to reconstruct frames with cell activities and at the same time detect and classify possible cell events. In Fig. 6, from epoch 1 to 20, the reconstructed frames at the end of the sequence are blurrier than the first frames because the further away from the time step $t$ the more difficult it is to predict the state of the cell population based on the learned dynamics $h_{t-1}$. At epoch 30, after the model is able to capture the complete dynamics of the cell population, the event detection branch, which outputs $y_{t:t+k}$, is then optimized and the network produces the complete reconstructed sequence with cell events, such as cell divisions. Our findings on model's behaviour show that our network models $P(y_{t:t+k} \mid x_{t:t+k})$ after $P(h_{t-1} \mid x_{1:t-1})$. This behaviour can be explained by the fact that the cross-entropy loss associated with the non-dividing cells' regular movements is greater than that of the dividing cells undergoing cell division, which is only a transient period in a cell's life cycle and lasts for a few frames only. Although the event detection result $y_{t:t+k}$ contains temporal and spatial information of cell events, the resolution is low due to the downscaling max-pooling step, as illustrated in Fig. 7. Therefore, in order to reconstruct cell division events occurring at the exact location of the mother cells, the model needs to define the future location of cells by learning their regular movements first. After epoch 40, the model learns to sharpen the reconstructed output to make it appear more visually similar to the original input frames.

In Fig. 8 we demonstrate the ability of our model to group visually similar events. Cell divisions are the most noticeable events with the largest variations in visual characteristics, as seen in Fig. 8(b). Other events are either cell movements that deviate from the usual motility captured by the Encoding ConvLSTM (E) and require further extracted information from Event Detection ConvLSTM (Ev) to reconstruct, or those events that do not seem to contain essential information (Fig. 8(c-f)). In our experiments, we heuristically set the number of class groups to 16. A higher number of class groups can be used because the neural network is flexible that it can turn off some artificial neurons of the Reconstruction part (R) by reducing their parameter values to ignore extra classes. A low number of class groups, for example 4 in our experiments, can result in forced grouping of multiple events into one class. From Fig. 8, some of the event classes are biologically important cell events, some may not depict specific biological activities but show cell behaviour that is different from the norm (e.g. cells protruding arms or moving faster than regularly seen), and some may be ignored by the model in reconstructing the input frames. However, prominent cell events, such as cell divisions, are always detected and grouped into one class, according to our experimental results.

In this work, we determined a small set of architecture hyper-parameters following the guidance of prior work on ConvLSTMs by Shi et al [35] and used Xavier initialization [42] to initialize the model's weights. However, these hyper-parameters can be further optimized or chosen differently depending on applications. Also, there are different weight initialization methods, such as data-dependent initializations, that might gain better results with faster convergence in the model training process [53-55]. Despite using a standard set of configuration, our unsupervised model showed high accuracy and its behaviours were analysed using the produced results. To further improve the results, an initial supervised screening process, such as one designed by Huh et al. [56]

to exclude easily recognizable non-cell division events and enhance cell event detection, can also be used to filter the input for our unsupervised model.

## 8. Conclusion

We developed an automated unsupervised method to detect and classify cell events by expanding a convolutional LSTM neural network with a branched structure for video data without annotations. Our method allows for the generation of two outputs in an unsupervised manner. Event detection information are extracted in the middle and the reconstructed frames at the end of the pipeline for comparison to the original training frames. We built an unsupervised LSTM model, and compared it against its supervised counterpart and published supervised HCRF methods. Our unsupervised model learned the dynamics of video cellular events and had results comparable to those from supervised methods. Furthermore, our models, though trained on one video, showed consistent and transferable performance across videos of different characteristics. We suggest that our unsupervised model has the potential to be used for the detection or identification of cell events in biological experiments. In future work we will explore using a different loss function, such as with generative adversarial networks [52], to increase the sensitivity of the event detection branch in our model. We will also investigate the application of our model to different cell types and cell events, such as cell deaths.

# Appendix

Table A.1: Hyper-parameters for our model

| Hyper-parameter | Value |
| --- | --- |
| ConvLSTM hidden states | 32 |
| ConvLSTM input-to-state kernel size | 5x5 |
| ConvLSTM state-to-state kernel size | 5x5 |
| ConvLSTM input size | 64x64 |
| CNN1 kernel size | 5x5 |
| CNN2 kernel size | 1x1 |
| Learning rate | $10^{-3}$ |
| Decay rate | 0.9 |
| Number of classes of events | 16 |
| Number of frames for the Encoding branch ($x_{1:t-1}$) | 5 |
| Number of frames for the Event Detection branch ($x_{t:t+k}$) | 10 |

Table A.2: Cell division counts of videos

| Video | Cell division count |
| --- | --- |
| F0001 | 456 |
| F0002 | 379 |
| F0003 | 319 |
| F0004 | 324 |
| F0005 | 245 |